# Safeguarding AI in Medical Imaging: Post-Hoc Out-of-Distribution Detection with Normalizing Flows

Dariush Lotfi, Mohammad-Ali Nikouei Mahani, Mohamad Koohi-Moghadam, Kyongtae Ty Bae

***Abstract*—In AI-driven medical imaging, the failure to detect out-of-distribution (OOD) data poses a severe risk to clinical reliability, potentially leading to critical diagnostic errors. Current OOD detection methods often demand impractical retraining or modifications to pre-trained models, hindering their adoption in regulated clinical environments. To address this challenge, we propose a post-hoc normalizing flow-based approach that seamlessly integrates with existing pre-trained models without altering their weights. We evaluate the approach on our in-house-curated MedOOD dataset, designed to capture clinically relevant distribution shifts, and on the MedMNIST benchmark. The proposed method achieves an AUROC of 84.61% on MedOOD, outperforming ViM (80.65%) and MDS (80.87%), and reaches 93.8% AUROC on MedMNIST, surpassing ViM (88.08%) and ReAct (87.05%). This combination of strong performance and post-hoc integration capability makes our approach a practical and effective safeguard for clinical imaging workflows.**

## I. Introduction

ARTIFICIAL intelligence (AI) and deep learning have transformed medical imaging, achieving high diagnostic accuracy and operational efficiency [1]. However, these advances expose a critical limitation: the inability of AI systems to recognize or quantify uncertainty in their predictions [2]. Real-world cases have shown that models can produce confident yet incorrect outputs when encountering unfamiliar inputs, leading to diagnostic errors with potentially serious consequences [3]. Ensuring the reliability of AI-assisted diagnostics therefore requires robust mechanisms to detect and manage predictive uncertainty in clinical environments.

A central challenge is the detection of out-of-distribution (OOD) data: inputs that differ substantially from those used to train the model [4]. In medical imaging, OOD data may arise from variations in patient demographics, imaging modalities, acquisition protocols, or from novel pathologies unseen during training. When faced with such inputs, AI models often produce unreliable predictions that can compromise diagnostic decisions [3]. Reliable OOD detection is thus essential for the safe deployment of AI systems in healthcare.

A wide range of OOD detection strategies have been proposed, including both post-hoc and training-time approaches. Early post-hoc methods such as MSP [5] and ODIN [6] relied on softmax probabilities and temperature scaling but suffered from overconfidence. Energy-based techniques improved robustness by using logit-based energy scores. SHE [7] introduced modern Hopfield networks to model in-distribution (ID) patterns using alignment-based energy scores. ViM [8] combined residual-based principal subspace features with logits, whereas feature-based methods such as MDS [9], used Mahalanobis distances from intermediate representations. Other works, including NCI [10], leveraged Neural Collapse phenomena by measuring alignment with class weight vectors, and Gram [11] characterized feature correlations across layers using Gram matrices. Recent activation-shaping methods like ReAct [12], DICE [13], and ASH [14] reduced spurious activations through clipping or pruning, while SCALE [15] refined this concept through sample-specific activation scaling to improve ID–OOD separability without reducing ID accuracy.

Despite these advances, many methods depend on retraining, regularization, or auxiliary data, requirements that are impractical in medical imaging, where pre-trained models are constrained by regulatory approval and limited computational resources. To address these challenges, we propose a post-hoc OOD detection framework (**Fig. 1**) based on normalizing flows, a class of probabilistic generative models capable of exact likelihood estimation and efficient sampling [16]. Our approach operates in the feature space of pre-trained models rather than pixel space, emphasizing semantically meaningful distinctions and avoiding the low-level sensitivity typical of pixel-based density estimation methods [17-19].

In addition, we introduce MedOOD, a curated benchmark dataset that simulates clinically relevant distributional shifts across population, modality, and transformation categories.

"This work was conducted in the JC STEM Lab of Innovative Medical Imaging Research funded by The Hong Kong Jockey Club Charities Trust." *(Co-corresponding authors: Kyongtae Ty Bae; Mohamad Koohi-Moghadam)*

Dariush Lotfi, Mohammad-Ali Nikouei Mahani, Mohamad Koohi-Moghadam, and Kyongtae Ty Bae are with the Department of Diagnostic Radiology, Li Ka Shing Faculty of Medicine, The University of Hong Kong, Pok Fu Lam, Hong Kong (e-mail: lotfi@hku.hk; nikoueim@uni-marburg.de; koohi@hku.hk; baekt@hku.hk)

Our codes are publicly available at https://github.com/dlotfi/MedOODFlow.

Using this benchmark, we demonstrate that our flow-based detector achieves consistent performance gains over prior methods and generalizes effectively across diverse imaging conditions.

The main contributions of this work are as follows:

- A post-hoc OOD detection approach using normalizing flows applied to feature embeddings, requiring no model retraining or architectural modification.
- Introduction of the MedOOD dataset for systematic evaluation of clinically realistic medical OOD detection.
- Comprehensive experiments on MedOOD and MedMNIST demonstrating superior OOD detection performance of our approach compared with state-of-the-art post-hoc methods.

## II. Materials and Methods

### A. Model Architecture

We propose a density (likelihood) estimation approach for OOD detection that operates on high-level feature representations rather than directly on input pixels. By modeling the probability distribution of features extracted from a base model, our method uses the semantic richness of learned features for more effective distinction between ID and OOD samples. This is desirable in practice, as it allows us to safeguard the base model from producing inaccurate or unreliable predictions on inputs that deviate significantly from its training distribution (**Fig. 1b**). Our model architecture is inspired by the Real Non-Volume Preserving (RealNVP) model [20], transforming input feature vectors extracted from the base model into a latent space through a series of invertible mappings (**Fig. 1c**). Specifically, it employs four masked affine coupling flows [20], each preceded by an ActNorm layer [21] for stabilization. Each coupling layer transforms a subset of input dimensions conditioned on others.

The coupling network within each flow block consists of two multilayer perceptrons (MLPs), for the scale and translation functions. The MLPs have a hidden layer of 1024 dimensions, enabling them to capture complex relationships in the feature space. Mathematically, the transformation in each coupling layer is defined as:

$$\begin{gathered} \mathbf{y}_{1:d} = \mathbf{x}_{1:d} \\ \mathbf{y}_{d+1:D} = \mathbf{x}_{d+1:D} \odot \exp\left(s(\mathbf{x}_{1:d})\right) + t(\mathbf{x}_{1:d}) \end{gathered} \tag{1}$$

where $\mathbf{x} \in \mathbb{R}^D$ is the input feature vector, $\mathbf{y}$ is the transformed output, $d$ is the index dividing the input dimensions, $s(\cdot)$ and $t(\cdot)$ are scale and translation functions modeled by MLPs, and $\odot$ denotes element-wise multiplication. This design ensures that the transformation is invertible and that the Jacobian determinant, required for likelihood computation, is easy to compute.

The normalizing flow model was trained to maximize the log-likelihood of the ID feature vectors, $\mathbf{x}$. The training objective is expressed as:

$$\begin{aligned} \mathcal{L} &= -\frac{1}{N}\sum_{i=1}^{N} \log p_{\mathbf{X}}\left(\mathbf{x}^{(i)}\right) \\ &= -\frac{1}{N}\sum_{i=1}^{N} \log p_{\mathbf{Z}}\left(\mathbf{z}^{(i)}\right) \\ &+ \sum_{k=1}^{K} \log\left|\det\left(\frac{\partial f_k^{-1}\left(\mathbf{x}^{(i)}\right)}{\partial \mathbf{x}}\right)\right| \end{aligned} \tag{2}$$

where $N$ is the number of training samples, $p_{\mathbf{X}}(\mathbf{x})$ is the density of the input data, $\mathbf{z}^{(i)} = f(\mathbf{x}^{(i)})$ is the transformed latent variable, $p_{\mathbf{Z}}(\mathbf{z})$ is the base distribution (standard normal), $K$ is the number of coupling layers, and $f_k^{-1}$ denotes the inverse transformation at layer $k$. The term involving the determinant of the Jacobian accounts for the change of variables during transformation. The overall architecture allowed us to model the probability density function of the ID features extracted from the base model.

### B. Model Implementation and Evaluation

A key aspect of our method is that training was unsupervised with respect to OOD data, relying solely on ID samples. This eliminates the need for OOD examples during training, reducing potential biases and improving generalizability across diverse medical imaging scenarios. Model selection used AUROC on designated validation sets, and the model with the highest AUROC was chosen. This procedure was applied consistently across all benchmark datasets (see **Section III**).

To evaluate our OOD detection approach, we used pre-trained classification models for medical image analysis, aiming to detect OOD data without modifying their weights or outputs and without requiring OOD samples during training of our proposed detector. For 2D datasets, we employed ResNet-18 [22], a standard architecture for image classification. This architecture consists of multiple convolutional blocks that progressively extract higher-level features. Instead of using only the final output features, we applied average pooling to the features from all hierarchical levels and concatenated them to form a comprehensive feature vector for OOD detection (**Fig. 1c**). For datasets involving 3D medical images, we used a 3D variant of the same residual network architecture [23].

Our method operates in a post-hoc manner, so throughout training of the OOD detection model the weights of the base classifiers were kept frozen, preserving their learned representations. The training process utilized the Adam optimizer with a learning rate of $1 \times 10^{-4}$ over 100 epochs, with model parameters initialized randomly. Our implementation was carried out in PyTorch, using the established OpenOOD framework [24-26] for OOD detection. All experiments were performed on a single NVIDIA RTX 6000 GPU.

## III. Benchmark and Curated Dataset

Our method was evaluated using two datasets, providing a comprehensive assessment across varied imaging modalities and tasks: MedOOD, which we curated to simulate clinically relevant distributional shifts, and MedMNIST v2 [27], a benchmark dataset offering standardized medical imaging data for machine learning research.

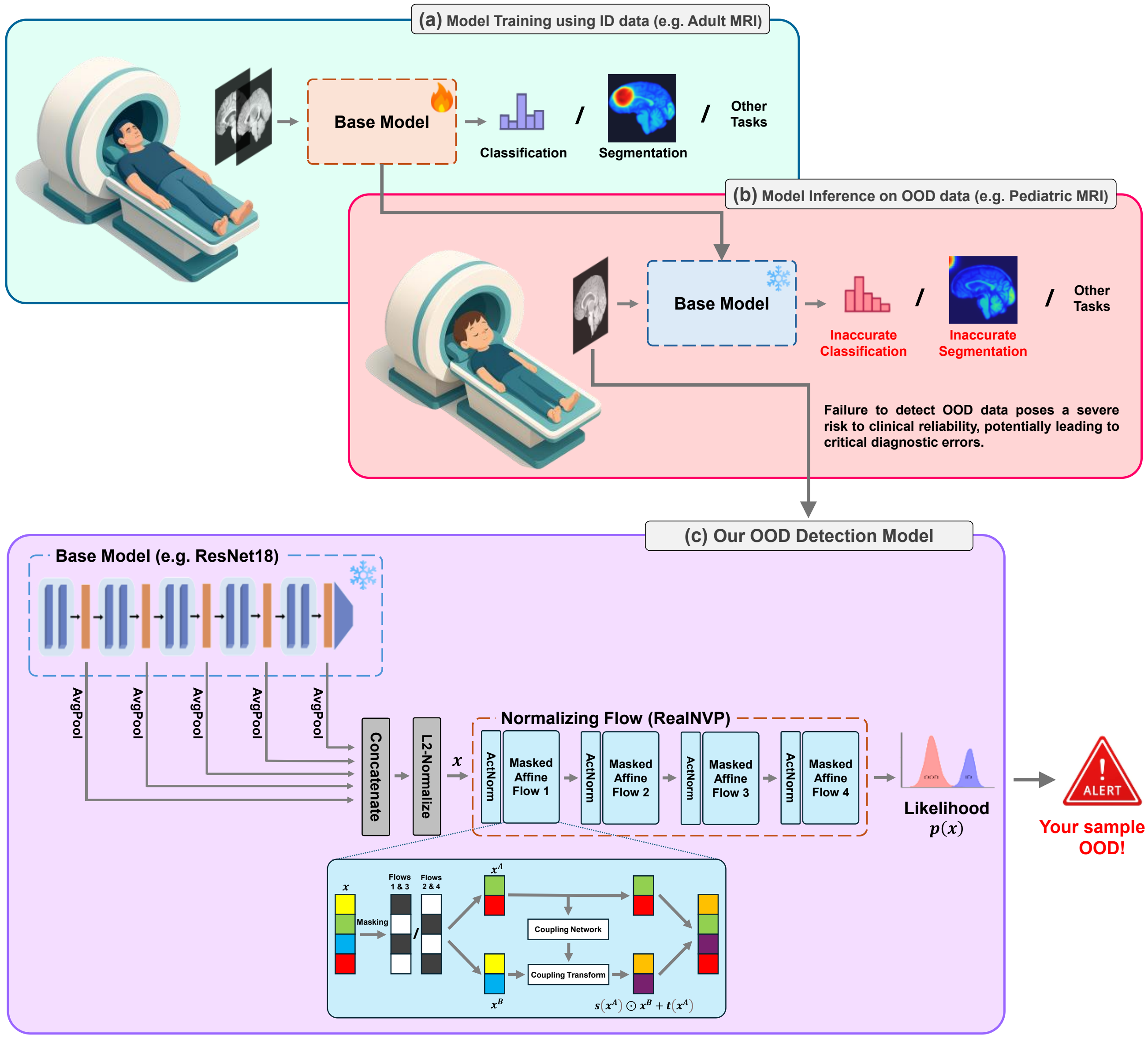

Fig. 1. Overview of the proposed method. (a) The base model is trained on ID data (e.g. Adult MRI) for AI tasks such as classification or segmentation. (b) In real-world scenarios, the base model may encounter OOD samples (e.g., Pediatric MRI) that result in high-confidence but inaccurate predictions. (c) Our method integrates seamlessly into clinical workflows without requiring model modifications or retraining. It aggregates and L2-normalizes base-model features, which are then processed by a RealNVP normalizing flow with four masked affine coupling layers to estimate likelihoods for distinguishing ID from OOD samples.

## A. MedOOD Curated Dataset

We curated a new OOD benchmark dataset, MedOOD, from publicly available medical imaging repositories, containing images with clinically relevant distributional shifts. The ID dataset consists of multi-center T1-weighted (T1w) brain MRIs of adult patients with glioma, selected from the BraTS 2020 [28-30] dataset, and used to train our base model for binary classification, distinguishing glioblastomas (GBM/HGG) from lower-grade gliomas (LGG). For OOD samples we built 21 OOD datasets, grouped into five categories [31, 32]. An overview and a few samples of the MedOOD datasets used in this study are provided in **Table I** and **Fig. 2**.

All brain imaging data underwent a preprocessing pipeline to ensure de-identification and consistency with the BraTS data. De-facing and skull-stripping was performed using the HD-BET algorithm [33]. The images were then registered to the SRI24 atlas [34] using ANTs [35], resampled to an isotropic voxel resolution of 1 mm³, standardized to an image size of 240×240×155 voxels, and normalized with intensity clipping at the upper and lower 0.1%. For non-brain datasets (abdominal and lumbar MRIs), we applied resampling to 1 mm³, normalization, and center cropping to match the target dimensions. For brain CT, we standardized intensities using a 40 HU (Hounsfield Unit) center and 100 HU width window.

TABLE I
OVERVIEW OF CURATED MEDOOD DATASETS USED IN THIS STUDY

| Category | Dataset | Description | # |
|---|---|---|---|
| **ID** | BraTS 2020 T1 [28-30] | Multi-center T1w brain MRIs of adults with glioma | 369* |
| **Control ID** | LUMIERE [36] | Pre-operative T1w brain MRIs of adults with gliomas from a different imaging center, simulating unseen clinical data | 80 |
| **Transformation Shifts** | Motion Artifact | Simulated head movements during acquisition by applying random rotations and translations | 75 |
| | Ghost Artifact | Emulated periodic motion such as cardiac or respiratory movement by replicating regions along axes | 75 |
| | Bias Artifact | Non-uniform illumination (bias field) generated via linear combination of polynomial basis functions | 75 |
| | Spike Artifact | Emulated Herringbone artifacts (periodic stripes) caused by RF pulse abnormalities due to aberrant k-space points | 75 |
| | Gaussian Noise | Random Gaussian noise | 75 |
| | Downsampling | Low-resolution or anisotropic voxels simulated by random downsampling and upsampling | 75 |
| | Scaling Perturbation | Altered brain size (halving or doubling) | 75 |
| | Gamma Alteration | Extreme contrast simulated by exponential intensity scaling | 75 |
| | Truncation | Missing data simulated by cropping half of the image in a random direction | 75 |
| | Erroneous Registration | Registration errors simulated by adding noise to the affine matrix | 75 |
| **Population Shifts** | BraTS 2023 Pediatric [37] | Pediatric GBM T1w MRIs aligned to adult templates | 99 |
| | BraTS 2023 Africa [38] | Adult MRIs from Sub-Saharan Africa, typically with lower quality and advanced disease due to late diagnosis | 60 |
| **Modality Shifts** | BraTS 2020 FLAIR [28-30] | FLAIR sequences of the BraTS 2020 test ID subjects | 75 |
| | BraTS 2020 T1ce [28-30] | T1ce (contrast-enhanced) sequences of the BraTS 2020 test ID subjects | 75 |
| | CQ500 [39] | Pre-contrast brain CT scans | 150 |
| **Diagnostic Shifts** | WHM 2017 [40] | T1w MRIs of patients with multiple sclerosis | 150 |
| | ATLAS R2.0 [41] | T1w MRIs of patients having brain stroke | 150 |
| | EPISURG [42] | T1w MRIs of epilepsy post-resection patients | 150 |
| | IXI [43] | T1w MRIs of young, healthy adults | 150 |
| **Organ Shifts** | CHAOS [44] | T1w abdominal MRIs | 80 |
| | Lumbar Spine [45] | Lumbar spine T1w MRIs | 150 |

* The ID dataset is divided into three subsets: training, validation, and testing, consisting of 274, 20, and 75 samples.

TABLE II
OVERVIEW OF MEDMNIST 2D BENCHMARK DATASETS USED IN THIS STUDY

| Category | Dataset | Description | # |
|---|---|---|---|
| **ID** | OrganAMNIST [46, 47] | Abdominal CT axial view | 58,850* |
| **Near OOD** | OrganCMNIST [46, 47] | Abdominal CT coronal view | 8,268 |
| | OrganSMNIST [46, 47] | Abdominal CT sagittal view | 8,829 |
| | ChestMNIST [48] | Chest X-ray of patients having various thoracic diseases | 22,433 |
| | PneumoniaMNIST [49] | Pediatric chest X-ray healthy or with pneumonia | 624 |
| **Far OOD** | PathMNIST [50] | Histological images of colorectal cancer tissues | 7,180 |
| | DermaMNIST [51] | Dermatoscopic images of common pigmented skin lesions | 2,005 |
| | RetinaMNIST [52] | Retina fundus images with varying diabetic retinopathy severity | 400 |
| | BloodMNIST [53] | Microscopic images of different blood cell types | 3,421 |

* The ID dataset is divided into three subsets: training, validation and testing, consisting of 34,561, 6,491, and 17,778 samples.

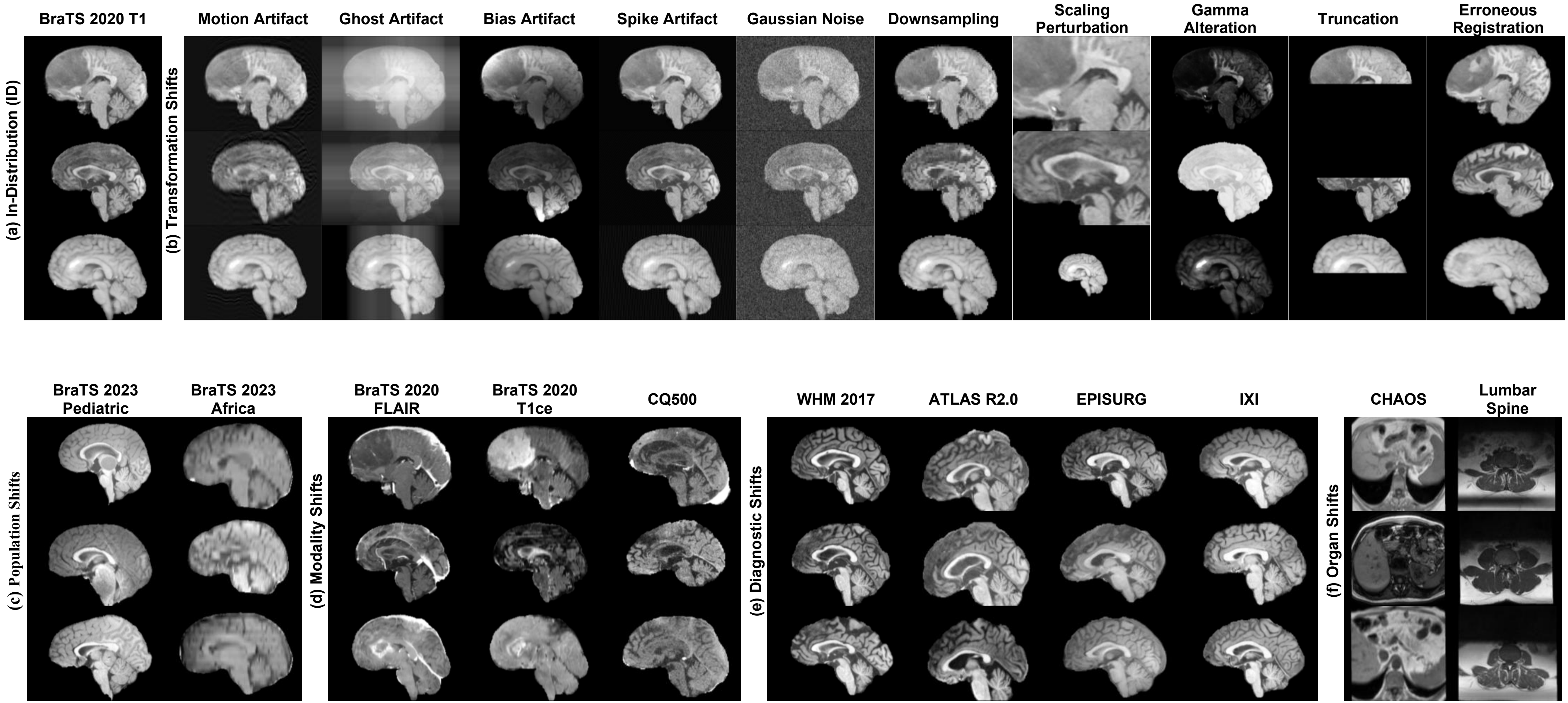

Fig. 2. A few samples of the datasets included in MedOOD. (a) In-Distribution (b) Transformation Shifts (c) Population Shifts (d) Modality Shifts (e) Diagnostic Shifts (f) Organ Shifts

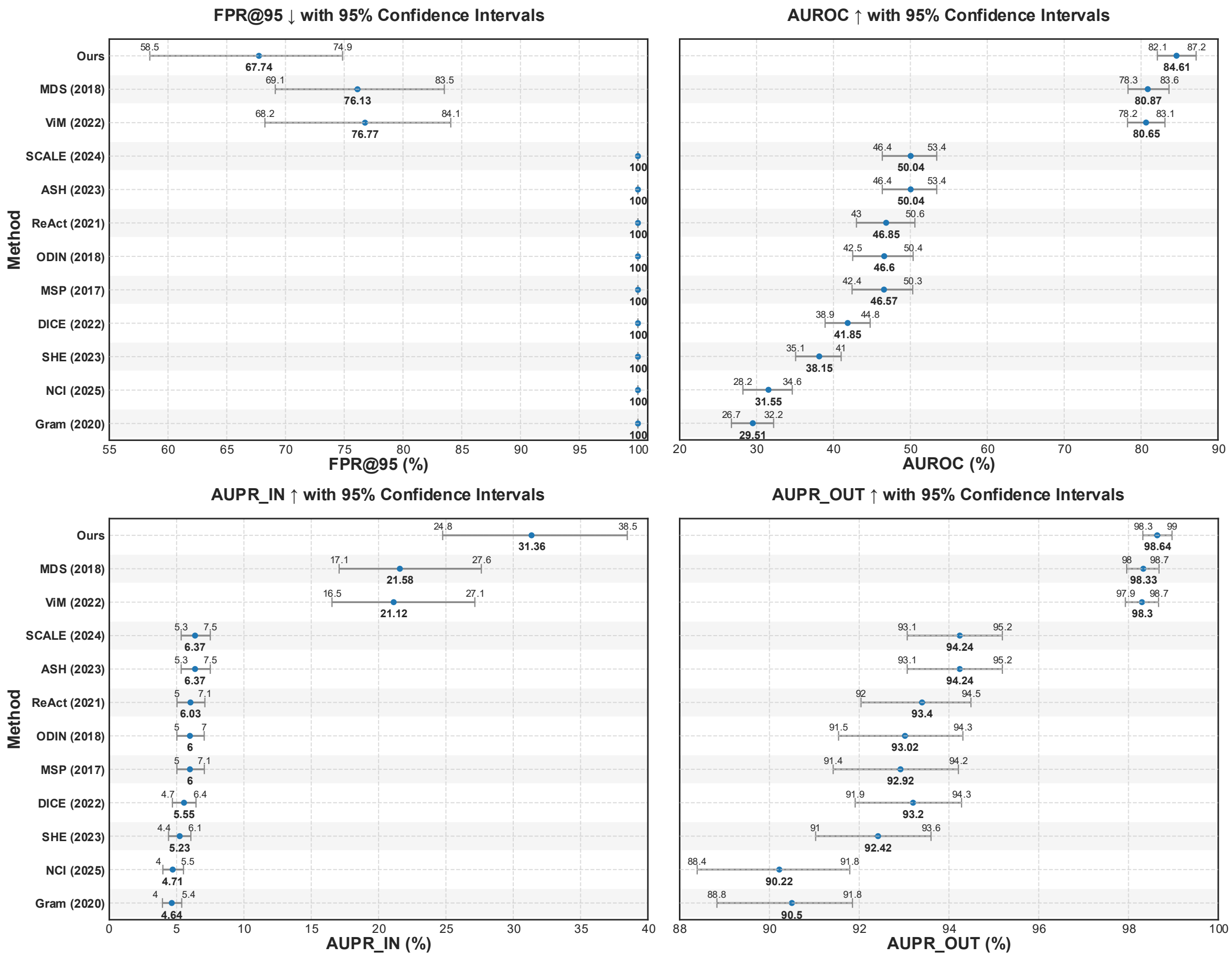

Fig. 3. Comparison of the OOD detection performance between our method and other post-hoc methods on MedOOD. Metrics were computed using micro-averaging.

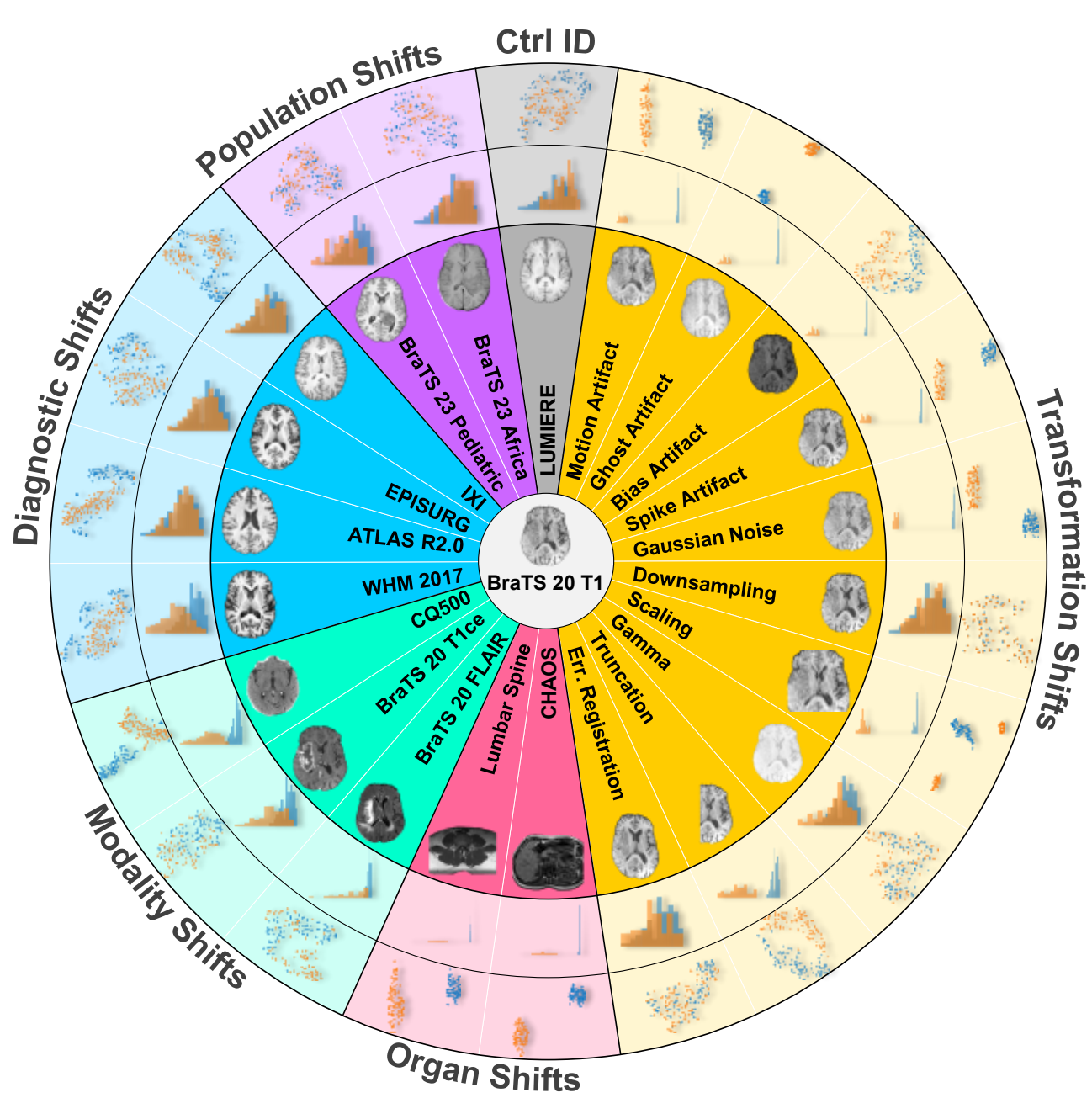

Fig. 4. OOD, ID and Control ID datasets in MedOOD, alongside log-likelihood histograms and t-SNE visualizations of features from the base model. The t-SNE visualization in feature space illustrates the degree of separation between the samples from the base model's perspective, while the histograms illustrate our model’s ability to assign distinct likelihood scores to ID versus OOD samples.

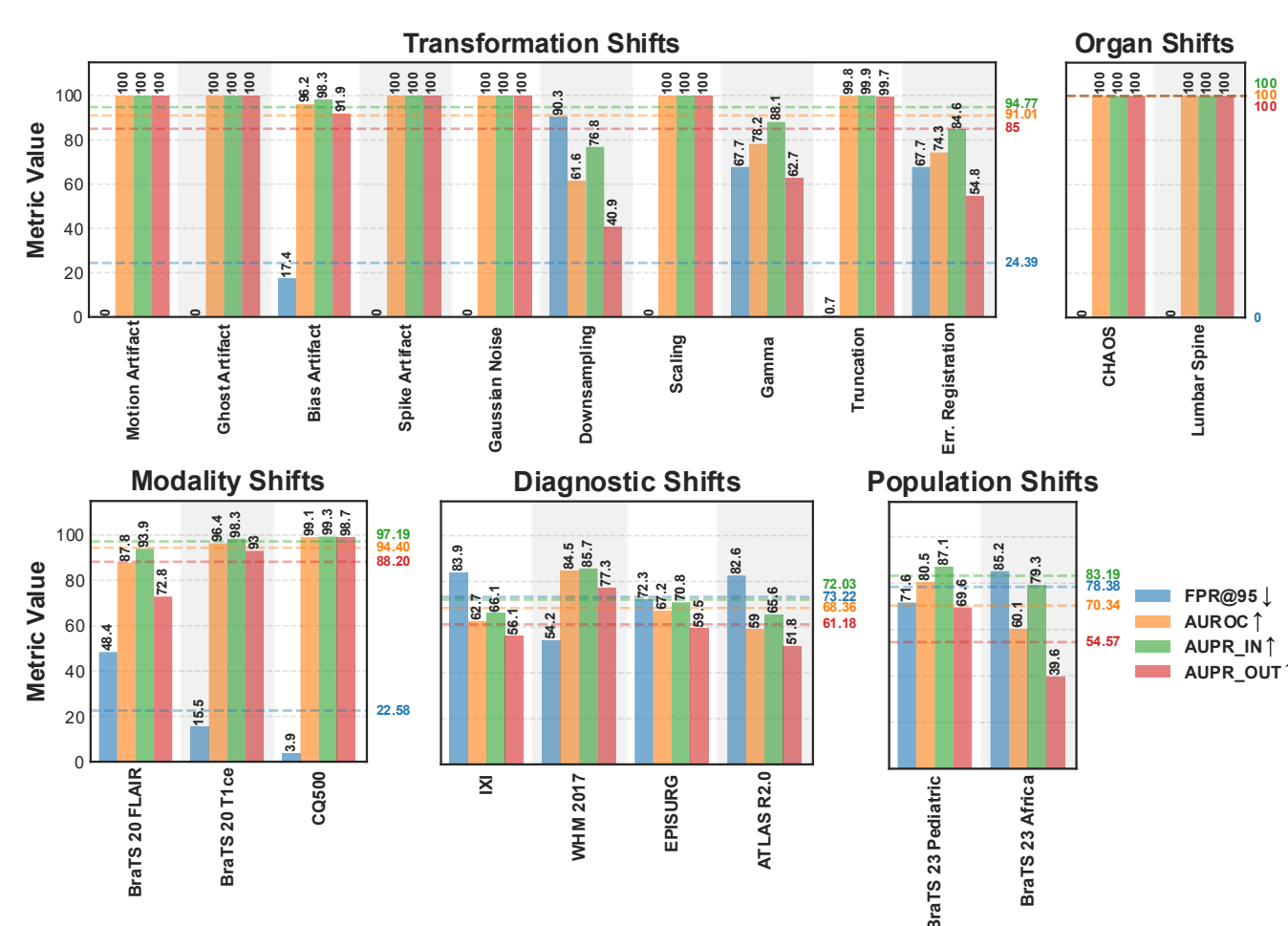

Fig. 5. OOD detection performance on different categories and datasets of MedOOD. The numbers on the right indicate averages per metric, matching the color of each metric's bars.

## B. MedMNIST Benchmark Datasets

MedMNIST v2 [27] covers a variety of medical imaging modalities, such as X-ray, computed tomography (CT), dermoscopy, microscopy, fundus photography, and histopathology. The datasets were pre-processed into a uniform size of 28×28 pixels for 2D images, facilitating rapid experimentation while preserving essential diagnostic features.

**Table II** provides an overview of the datasets included in MedMNIST 2D benchmark and used in our study. We chose OrganAMNIST [46, 47], consisting of axial CT images of abdominal organs in 11 classes, as ID dataset to train our base model for multi-class classification.

## IV. Evaluation and Statistical Analysis

We evaluated the performance of our approach using four standard OOD detection metrics [24]: False Positive Rate at 95% True Positive Rate (FPR@95), Area Under the Receiver Operating Characteristic Curve (AUROC), Area Under the Precision-Recall Curve for In-Distribution Samples (AUPR_IN), and Area Under the Precision-Recall Curve for Out-of-Distribution Samples (AUPR_OUT). They collectively measure specificity, threshold-independent performance, and the precision-recall trade-off for both ID and OOD samples.

To compare our method against baseline OOD detection approaches, we conducted a series of statistical analyses. First, we computed all four metrics, and their corresponding 95% confidence intervals (CIs) using the percentile bootstrap method with 1,000 iterations to quantify uncertainty. Second, to assess statistical significance, we conducted bootstrapping tests for AUROC differences with a significance threshold of $P < 0.05$. Finally, paired AUROC comparisons were performed using the DeLong method.

### A. Performance on the MedOOD Curated Dataset

We first evaluated the effectiveness of our OOD detection method on the curated MedOOD dataset. As shown in **Fig. 3**, our method achieved an AUROC of 84.61%, outperforming state-of-the-art methods such as MDS [9] (80.87%) and ViM [8] (80.65%). For FPR@95, our model achieved a value of 67.74%, showing better specificity compared to MDS (76.13%) and ViM (76.77%). To assess the statistical significance of these improvements, we performed a comparative analysis using both DeLong's test and bootstrap resampling (1,000 iterations). The AUROC improvement over ViM (95% CI: 78.24–83.11) was statistically significant: DeLong's $P = 0.0105$, and bootstrap $P = 0.006$, as well as the improvement over MDS (95% CI: 78.30–83.63): DeLong's $P = 0.0126$, and bootstrap $P = 0.01$.

### B. Evaluating Robustness Across Distributional Shift Categories

We further validated the robustness of our method on the curated MedOOD dataset by assessing performance across different categories, including transformation, population, modality, diagnostic, and organ shifts. For transformation shifts, the model demonstrated strong robustness, achieving near-perfect AUROC and low FPR@95 on most simulated artifacts (e.g., motion, ghost, spike, and noise). This suggests that simulated artifacts often result in feature distributions that are well separated from the ID data. However, more subtle transformations, such as downsampling and gamma alterations, were harder to detect, reflected in greater overlap in log-likelihood distributions and t-SNE projections (**Fig. 4**). This highlights the difficulty of identifying OOD samples that closely resemble ID data in low-level statistics.

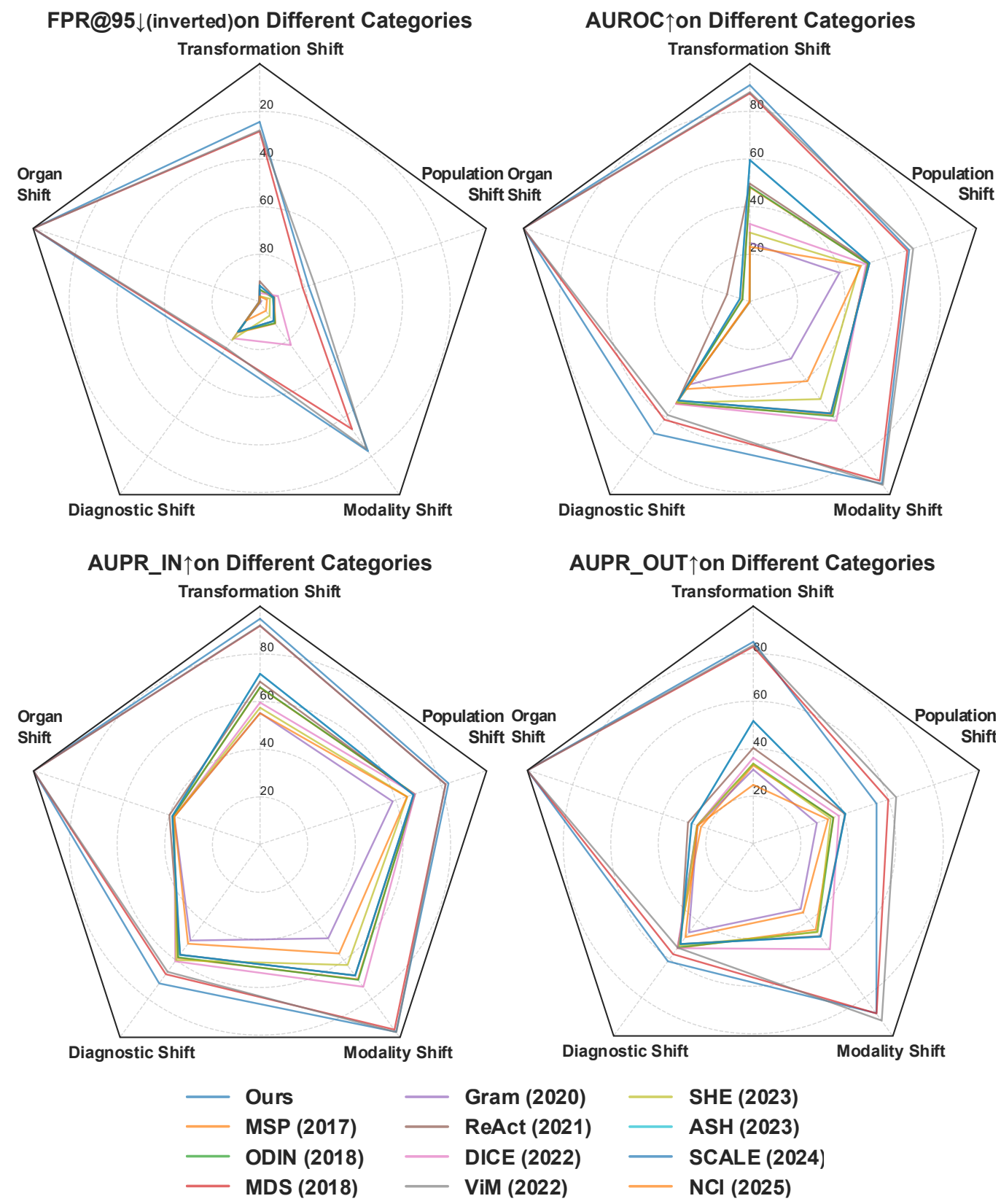

Fig. 6. Comparison of the OOD detection performance between our method and other post-hoc methods on different categories of MedOOD. The axis of FPR@95 is inverted to improve visualization.

Population shifts, such as the transition to pediatric or geographically distinct cohorts, presented another layer of complexity. Here, the model's performance decreased, with a drop in AUROC and AUPR_OUT, particularly for the BraTS 2023 Africa dataset. This suggests that demographic and clinical heterogeneity may result in OOD samples that are harder to distinguish from ID cases, likely due to underlying similarities in imaging protocols or disease manifestation. For modality shifts, the model exhibited consistently high generalizability. In contrast, diagnostic shifts yielded more variable results: while some conditions (e.g., multiple sclerosis in WHM 2017) were well detected as OOD, others (e.g., stroke in ATLAS R2.0) were less clearly separated, possibly reflecting overlapping imaging features with the ID cohort.

As expected, the model achieved perfect discrimination for organ shifts, consistently identifying non-brain MRIs as OOD. This outcome served as a sanity check, confirming that the method could easily recognize cases with substantial anatomical differences from the training distribution (**Fig. 5**). To contextualize our method's performance, we compared it with state-of-the-art post-hoc OOD detectors across all shift categories. **Fig. 6** presents a radar plot summarizing average performance using FPR@95 (lower is better), AUROC, AUPR_IN, and AUPR_OUT (higher is better). Our method consistently outperformed or matched existing baselines, particularly excelling in transformation and diagnostic shifts.

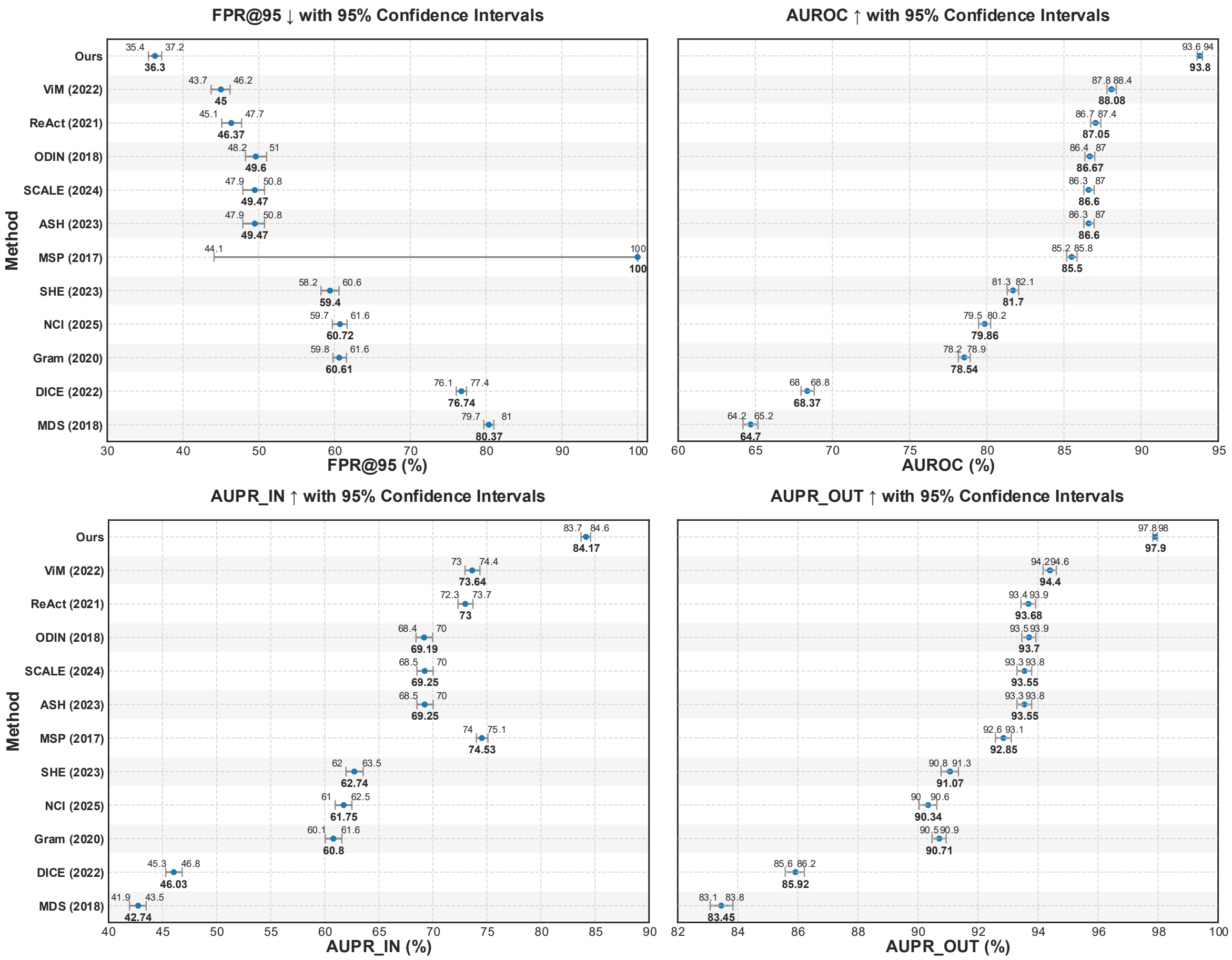

Fig. 7. Comparison of the OOD detection performance between our method and other post-hoc methods on MedMNIST. Metrics were computed using micro-averaging.

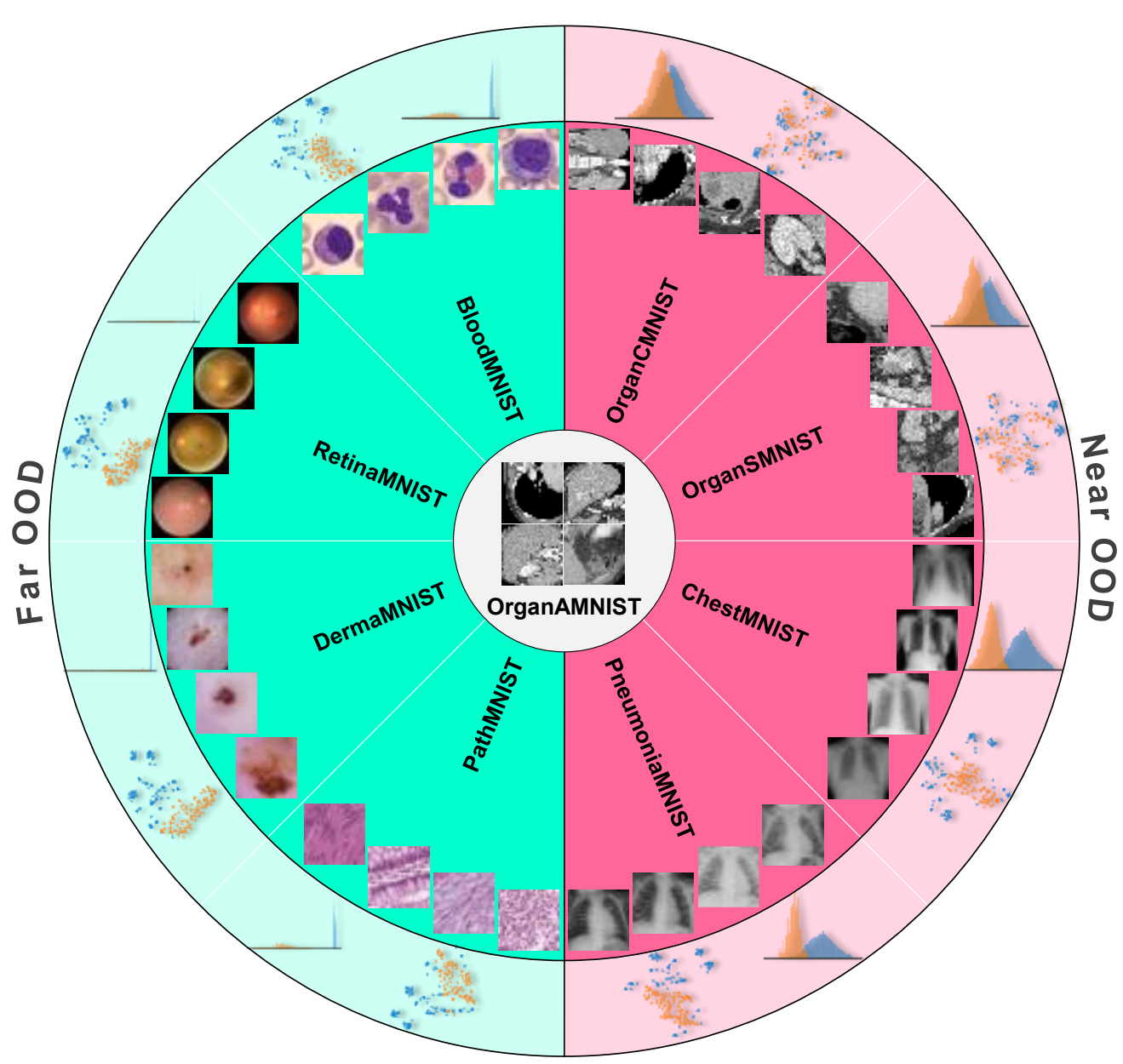

Fig. 8. OOD and ID datasets in MedMNIST, alongside log-likelihood histograms and t-SNE visualizations of features from the base model. The t-SNE visualization in feature space illustrates the degree of separation between the samples from the base model's perspective, while the histograms illustrate our model’s ability to assign distinct likelihood scores to ID versus OOD samples.

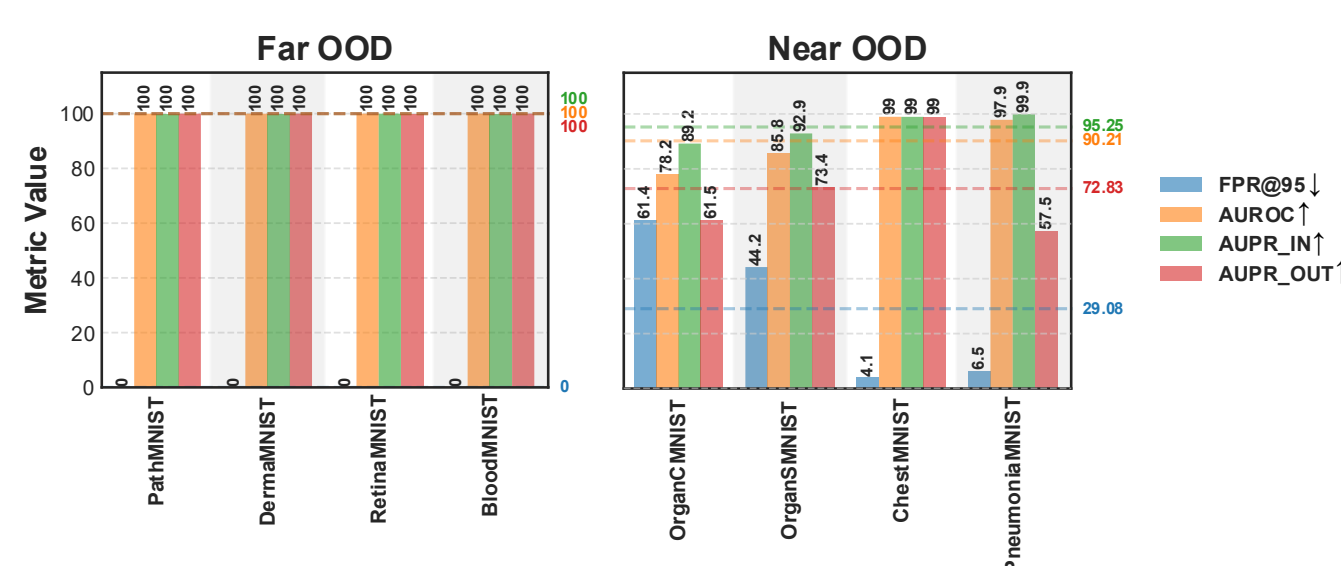

Fig. 9. OOD detection performance on different categories and datasets of MedMNIST. The numbers on the right indicate averages per metric, matching the color of each metric's bars.

### C. Performance on the MedMNIST Benchmark Dataset

We further evaluated our method on the MedMNIST benchmark against eleven publicly available post-hoc OOD detectors using the same metrics and comparison framework as in MedOOD evaluation. Our method achieved the best performance across all metrics (**Fig. 7**), with an AUROC of 93.8%, significantly outperforming state-of-the-art methods such as ViM [8] (88.08%) and ReAct [12] (87.05%). It also attained the highest AUPR_IN (84.17% vs. 73.64% for ViM and 73% for ReAct) and AUPR_OUT (97.9% vs. 94.4% for ViM and 93.68% for ReAct). Our model's improvement over the best baseline, ViM, was statistically significant. AUROC

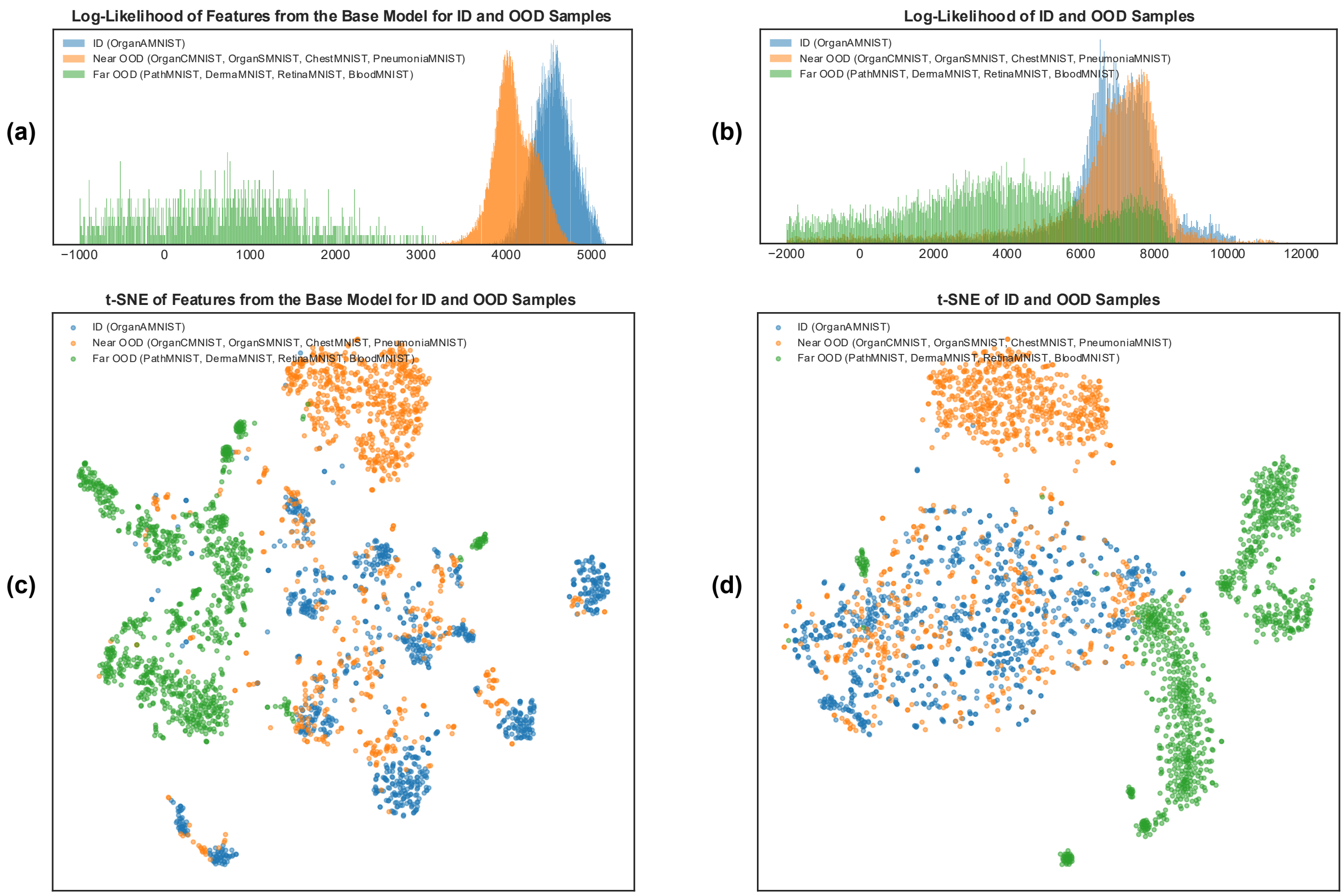

Fig. 10. Comparative analysis of feature space and pixel space for OOD detection on MedMNIST. (a) Predicted log-likelihood histogram of features extracted from the base model for ID and near/far OOD samples, where greater separability indicates improved OOD detection. (b) Predicted log-likelihood histogram of ID and near/far OOD samples in pixel space, showing increased overlap compared to feature space. (c) t-SNE visualization of the base model's feature space, illustrating clear ID clusters and separation of OOD samples. (d) t-SNE visualization of the samples in pixel space, revealing greater entanglement compared to feature space.

comparisons using both DeLong's test and bootstrap analysis yielded $P < 0.001$.

### D. Evaluating Robustness Across Near and Far OOD Categories in MedMNIST

To further assess the generalizability of our approach, we evaluated its OOD detection performance separately on datasets in both near and far OOD categories within the MedMNIST (**Fig. 8** and **9**). Histograms and feature visualizations showed clear trends: far OOD datasets (e.g., PathMNIST, BloodMNIST) displayed strong separation from ID samples, whereas near OOD datasets (e.g., OrganCMNIST, OrganSMNIST) showed greater overlap due to their similarity to the ID distribution (**Fig. 10a**).

Our model demonstrated robust performance across both shift types. For far OOD datasets, those involving distinct imaging modalities and anatomical regions, the model achieved perfect discrimination across all metrics, as expected given their substantial differences from the abdominal CT ID data. Near OOD datasets, including alternative CT views and chest X-rays (**Fig. 10b**), were more challenging but still well handled. OrganCMNIST and OrganSMNIST showed moderate FPR@95 and lower AUPR_OUT, reflecting their close resemblance to ID images; however, datasets such as ChestMNIST and PneumoniaMNIST remained clearly separable, indicating the model's ability to detect subtle distributional shifts.

### E. Evaluating OOD Performance in Feature Space vs. Pixel Space

Analysis of predicted log-likelihoods in feature space showed strong separability between ID and OOD samples on MedMNIST. As shown in **Fig. 10a**, far OOD samples were distinctly shifted toward lower likelihood values, while near OOD samples fell between the ID and far OOD. This gradation reflected the degree of distributional shift, with the model assigning lower confidence to samples that were more semantically or visually distinct from the training distribution. The t-SNE visualization of the base model's features (**Fig. 10c**) reinforced this, revealing well-defined clusters for the eleven OrganAMNIST ID classes and distinct positioning of OOD samples, with far OOD data showing the greatest separation. These results indicate that the base model's feature space captures meaningful semantic structure, enabling the normalizing flow to effectively distinguish ID from OOD data. In contrast, applying the normalizing flow directly in pixel space led to substantial performance degradation. Log-likelihood distributions of ID and OOD samples overlapped heavily (**Fig. 10b**), especially for near OOD datasets, highlighting the failure of pixel space density estimation to reliably differentiate between in- and out-of-distribution data. The corresponding t-SNE plot (**Fig. 10d**) further confirmed this, revealing a highly entangled representation of ID and OOD samples with no clear separation or clustering. This suggests that pixel space representations lacked the semantic abstraction necessary for robust OOD detection. Quantitatively, AUROC dropped from 93.8% (feature space) to 70.59% (pixel space), FPR@95 rose from 36.3% to 89.82%, and both AUPR_IN and AUPR_OUT were substantially lower in pixel space (38.72% vs. 84.17% and 87.83% vs. 97.9%). These findings align with prior research [17-19], which showed that deep generative models in pixel space often assigned high likelihoods to OOD data, particularly for semantically similar or structurally simple samples.

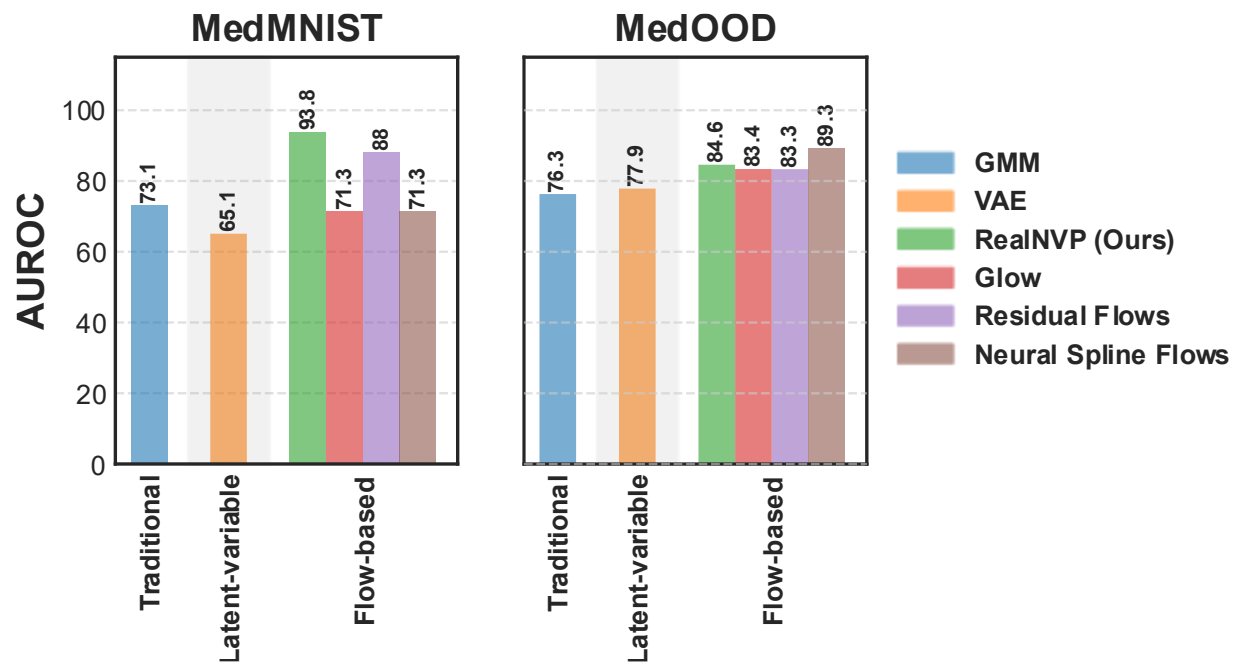

Fig. 11. AUROC for density-estimation-based OOD detection methods. Methods are grouped as Traditional, Latent-variable, and Flow-based. AUROC values are micro-averaged.

### *F. Comparison Across Density Estimators*

We evaluated additional density-estimation-based models on the same hierarchical feature representations used in this study. The compared families included a traditional mixture model (GMM), a latent-variable model (VAE) [54], and three flow-based models beyond RealNVP (Glow [21], Residual Flows [55], and Neural Spline Flows [56]). All models were trained and evaluated under identical settings for a fair comparison.

Although VAEs do not yield exact likelihoods, the negative evidence lower bound (ELBO) provides a tractable scoring surrogate that serves as a consistent, monotonic proxy for density.

On MedMNIST, our RealNVP-based estimator achieves the highest AUROC among the evaluated models. On MedOOD, RealNVP remains competitive with newer flows, with Neural Spline Flows performing slightly better, while all flow-based models match or exceed the GMM and VAE baselines (**Fig. 11**).

These results indicate that the proposed model effectively captures complex feature-space distributions while maintaining architectural simplicity. Moreover, the reasonable performance of both traditional and latent-variable models in feature space supports the discussion in **Section IV-E**, reinforcing that feature-space density estimation provides a reliable basis for OOD detection.

TABLE III
INFERENCE TIME AND OFFLINE COSTS

| Method | Inference Time (ms) | Offline Cost (s) |
|---|---|---|
| Ours | 17.9 | 128* |
| MSP [5] | 0.6 | - |
| ODIN [6] | 2.9 | 109† |
| MDS [9] | 16 | - |
| Gram [11] | 30.6 | 46†§ |
| ReAct [12] | 15.5 | 65†§ |
| DICE [13] | 16.3 | 17§ |
| ViM [8] | 15.8 | 53†§ |
| SHE [7] | 0.4 | 19§ |
| ASH [14] | 15.7 | 94† |
| SCALE [15] | 15.8 | 96† |
| NCI [10] | 0.5 | 78†§ |

* Training the flow model on the 274 ID training samples for 100 epochs. † Hyperparameter selection via grid search using AUROC on the ID and OOD validation sets (20 samples each). § One-time setup (e.g., computing ID statistics) performed once.

## V. COMPLEXITY ANALYSIS

We report per-sample inference latency for our approach and all baseline post-hoc OOD detectors under a unified setup. All methods use the same base classifier (ResNet3D-18) and the same inputs from MedOOD. All timings include the forward pass of the base model as well as the method-specific post-hoc scoring, and were measured on a single NVIDIA RTX 6000 GPU. **Table III** lists the inference time per sample and one-time/offline costs (when applicable), such as flow training or grid-search calibration.

Our flow-based method has comparable per-sample latency to common baselines and is faster than computationally intensive approaches such as Gram [11], while being slower than ultra-light score-based methods such as MSP [5]. Its offline cost is a one-time training stage, similar in magnitude to the calibration or statistic-computation steps required by several baselines.

## VI. ABLATION STUDY

To further investigate the factors contributing to the performance of our proposed normalizing flow-based method, we conducted a comprehensive ablation study.

### *A. Effect of Aggregating Hierarchical Features*

We analyzed the effect of aggregating features from different numbers of hierarchical levels across multiple base models by concatenating the last $k$ stages ($k \in \{1, \cdots, 5\}$) taken at natural stage boundaries: after stride-2 blocks in ResNet/ResNet3D [22, 23], transition pools in DenseNet3D [57], and patch-merging in SwinTransformer3D [58]. We then trained the same normalizing flow detector on the resulting representations.

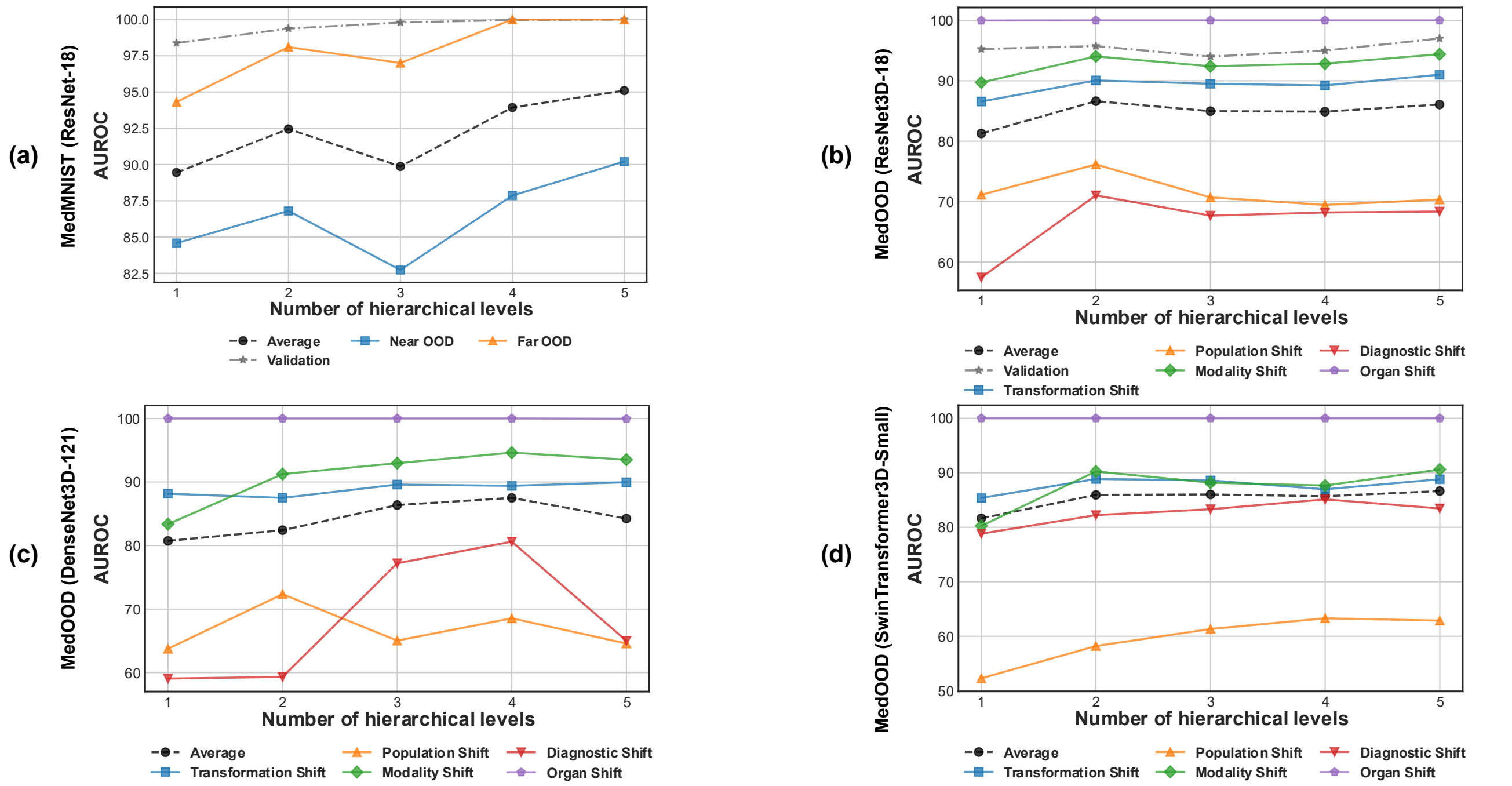

Fig. 12. Effect of aggregating hierarchical feature levels on OOD detection performance. (a) ResNet-18 on MedMNIST (b) ResNet3D-18 on MedOOD (c) DenseNet3D-121 on MedOOD (d) SwinTransformer3D-Small on MedOOD

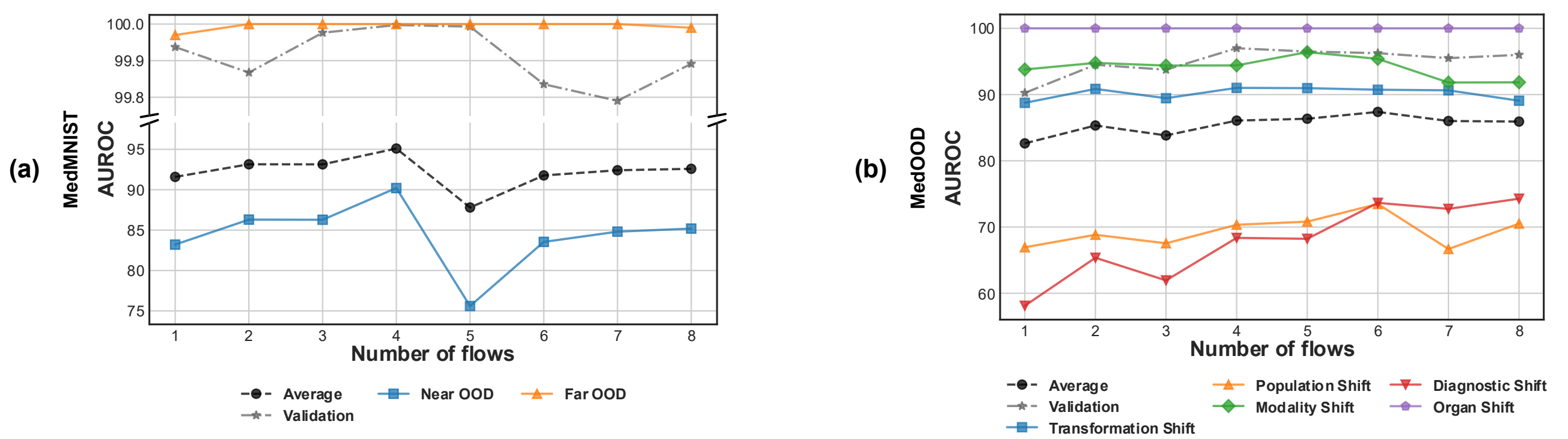

Fig. 13. Effect of the number of flow blocks on OOD detection performance of the normalizing flow. (a) MedMNIST (b) MedOOD

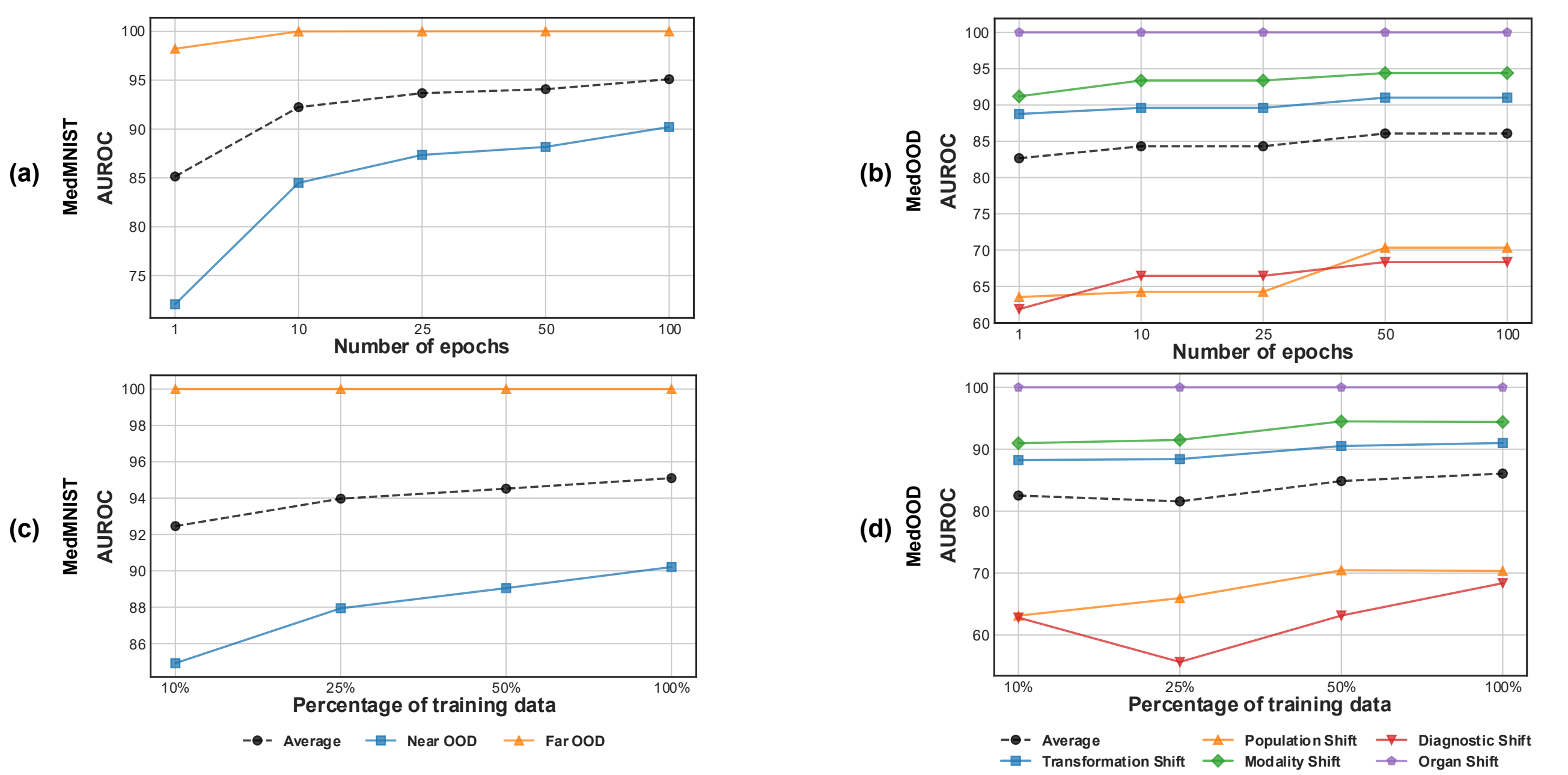

Fig. 14. Effect of training duration and data availability on OOD detection performance. (a, b) Impact of the number of epochs on performance for MedMNIST and MedOOD. (c, d) Impact of using different percentages of training data on performance for MedMNIST and MedOOD.

Across both MedOOD and MedMNIST, validation AUROC generally improved with larger $k$ and was maximal at $k = 5$ (**Fig. 12a** and **12b**); we therefore adopt $k = 5$ as the default to avoid architecture-specific tuning. On the test sets, using only the final level ($k = 1$) underperformed multi-level representations. For ResNet3D the average AUROC peaked at $k = 2$ but was only marginally higher than at $k = 5$ (**Fig. 12b**), and DenseNet3D and SwinTransformer3D showed monotonic or near-monotonic gains up to $k = 5$ (**Fig. 12c** and **12d**). Split-wise variations exist, for example, the Diagnostic split with ResNet3D attains its best value at $k = 2$, but combining $\geq 3$ stages consistently avoids the worst cases, and $k = 5$ is never substantially worse than the best $k$ for any split. Taken together, these results indicate that aggregating hierarchical features improves robustness while remaining simple to configure.

Observing the same qualitative pattern across distinct base model architectures (ResNet-18, ResNet3D-18, DenseNet3D-121, and SwinTransformer3D-Small) demonstrates that the proposed post-hoc normalizing-flow OOD detection method is robust to the choice of base model and transfers across both CNN and Transformer backbones without modification.

### *B. Effect of the Number of Flow Blocks*

We investigated the depth of the normalizing flow by varying the number of coupling blocks from 1 to 8 while keeping all other settings fixed. As shown in **Fig. 13**, validation AUROC increases with depth and reaches its maximum at 4 blocks. On the test sets, MedMNIST attains its highest average AUROC at 4 blocks; adding more blocks yields no consistent gains. On MedOOD, the average AUROC exhibits modest, mostly monotonic improvements up to 6 blocks, after which performance saturates. Across splits, the easiest categories remain near-ceiling regardless of depth, whereas the harder ones (e.g., Population/Diagnostic) benefit from increasing depth up to a certain level. Considering generalizability and simplicity, we adopt four flow blocks as the default in the main experiments.

### *C. Effect of Training Duration and Data Availability*

We explored the effect of training duration by varying the number of training epochs. Among all configurations tested, training the model for 100 epochs yielded the optimal performance, achieving the highest average AUROC values (95.1% for MedMNIST and 86.07% for MedOOD). However, training for only 10 epochs achieved 92.24% average AUROC on MedMNIST and 84.31% on MedOOD, corresponding to approximately 97% of the performance at 100 epochs (**Fig. 14a**). This demonstrates that the model achieved most of its performance gains early in training, making shorter training durations a highly efficient option. Additionally, to evaluate the robustness of our method under limited data settings, we trained the model on varying percentages of the training data (25%, 50%, and 100%). Our method demonstrated strong performance even when trained on only 25% or 50% of the available data, achieving average AUROC values of 94.52% on MedMNIST and 84.86% on MedOOD at 50% of the data, compared to 95.1% and 86.07% with full data, respectively (**Fig. 14b**). This underscores the data efficiency of our approach, which can maintain robust OOD detection performance even under resource-constrained conditions.

## VII. Conclusion

We presented a post-hoc OOD detection method for medical imaging based on normalizing flows, achieving superior performance to prior approaches on both the MedOOD and MedMNIST benchmarks.

Unlike conventional density-estimation approaches that operate in pixel space, our method estimates likelihoods in the feature space of pre-trained models. Feature-space modeling captures semantically meaningful representations and is less affected by low-level image statistics, improving separation between ID and OOD samples. Moreover, unlike OOD methods that rely solely on softmax probabilities or logit-based scores, our approach evaluates likelihoods independently of task-specific outputs. This design theoretically allows it to generalize across models trained for different objectives, including classification, segmentation, and regression, broadening its applicability in medical imaging.

A further advantage of the framework is its post-hoc design, which eliminates the need to retrain or modify pre-trained model weights. This property directly supports clinical AI safety, where regulatory and computational constraints often restrict alterations to established AI systems. In deployment, the OOD detector can function as an auxiliary module within existing PACS or AI pipelines. During inference, feature embeddings are processed by the trained normalizing flow to yield an OOD score for each study. This score can be stored as DICOM metadata and displayed in the PACS viewer or AI dashboard. When it exceeds a predefined threshold, the system can automatically flag the study, suppress automated outputs, or trigger a quality-control alert.

We also introduced MedOOD, an in-house dataset of 21 OOD sets spanning five clinically meaningful shift categories. It provides a standardized benchmark for evaluating OOD methods under realistic demographic, modality, and acquisition variations. By releasing both the dataset and implementation, we aim to foster reproducibility and further research on AI reliability in healthcare.

Despite strong overall performance, several limitations remain. Detection accuracy varied across shift types, with challenges in subtle transformations (e.g., downsampling) and population shifts (e.g., geographically or demographically distinct cohorts) where OOD samples closely resemble in-distribution data. These subtle transformations often preserve high-level semantic features, meaning that feature-space representations change little even when pixel-level statistics shift, making them inherently difficult for likelihood-based models to detect. Although the method performed well on most diagnostic shifts, variability across datasets suggests room for improvement in capturing fine-grained differences linked to unseen or rare pathologies. Future work should enhance robustness to subtle appearance changes and population-level variations through improved feature-space modeling, including multi-scale or texture-aware feature representations, more precise calibration of likelihood thresholds across domains, and the exploration of typicality-based testing [59] to complement likelihood estimation.

Finally, the transformation shifts in MedOOD rely on simulated artefacts that, while physically grounded, cannot fully capture the complexity and variability of real clinical distribution shifts. Factors such as scanner calibration, patient physiology, acquisition settings, and site-specific workflows jointly influence image appearance in ways that are difficult to reproduce synthetically. Nevertheless, these controlled perturbations serve as valuable stress tests for assessing OOD robustness under reproducible and interpretable conditions. Future versions of MedOOD will aim to narrow this realism gap through scanner-informed k-space modeling and expert Turing-test evaluations, where radiologists assess the perceptual plausibility of simulated artefacts.

## References

[1] P. Rajpurkar, E. Chen, O. Banerjee, and E. J. Topol, "AI in health and medicine," *Nature Medicine 2022 28:1,* vol. 28, no. 1, 2022-01-20, doi: 10.1038/s41591-021-01614-0.

[2] L. Huang, S. Ruan, Y. Xing, and M. Feng, "A review of uncertainty quantification in medical image analysis: Probabilistic and non-probabilistic methods," *Medical Image Analysis,* vol. 97, 2024/10/01, doi: 10.1016/j.media.2024.103223.

[3] Y. Yang, H. Zhang, J. W. Gichoya, D. Katabi, and M. Ghassemi, "The limits of fair medical imaging AI in real-world generalization," *Nature Medicine 2024 30:10,* vol. 30, no. 10, 2024-06-28, doi: 10.1038/s41591-024-03113-4.

[4] O. Zhang, J. B. Delbrouck, and D. L. Rubin, "Out of Distribution Detection for Medical Images," *Lecture Notes in Computer Science (including subseries Lecture Notes in Artificial Intelligence and Lecture Notes in Bioinformatics),* vol. 12959 LNCS, pp. 102-111, 2021, doi: 10.1007/978-3-030-87735-4_10.

[5] D. Hendrycks and K. Gimpel, "A Baseline for Detecting Misclassified and Out-of-Distribution Examples in Neural Networks," in *International Conference on Learning Representations*, 2017. [Online]. Available: https://openreview.net/forum?id=Hkg4TI9xl.

[6] S. Liang, Y. Li, and R. Srikant, "Enhancing The Reliability of Out-of-distribution Image Detection in Neural Networks," in *International Conference on Learning Representations*, 2018. [Online]. Available: https://openreview.net/forum?id=H1VGkIxRZ.

[7] J. Zhang *et al.*, "Out-of-Distribution Detection based on In-Distribution Data Patterns Memorization with Modern Hopfield Energy," in *The Eleventh International Conference on Learning Representations*, 2023. [Online]. Available: https://openreview.net/forum?id=KkazG4lgKL.

[8] H. Wang, Z. Li, L. Feng, and W. Zhang, "ViM: Out-Of-Distribution with Virtual-logit Matching," in *Proceedings of the IEEE Computer Society Conference on Computer Vision and Pattern Recognition*, 2022, vol. 2022-June: IEEE Computer Society, pp. 4911-4920, doi: 10.1109/CVPR52688.2022.00487.

[9] K. Lee, K. Lee, H. Lee, and J. Shin, "A Simple Unified Framework for Detecting Out-of-Distribution Samples and Adversarial Attacks," in *Advances in Neural Information Processing Systems*, S. Bengio, H. Wallach, H. Larochelle, K. Grauman, N. Cesa-Bianchi, and R. Garnett, Eds., 2018, vol. 31: Curran Associates, Inc. [Online]. Available: https://proceedings.neurips.cc/paper_files/paper/2018/file/abdeb6f575ac5c6676b747bca8d09cc2-Paper.pdf.

[10] L. Liu and Y. Qin, "Detecting Out-of-Distribution Through the Lens of Neural Collapse," in *Proceedings of the IEEE Computer Society Conference on Computer Vision and Pattern Recognition*, 2025, vol. in press: IEEE Computer Society, doi: 10.48550/arXiv.2311.01479.

[11] C. S. Sastry and S. Oore, "Detecting Out-of-Distribution Examples with Gram Matrices," in *Proceedings of the 37th International Conference on Machine Learning*, H. D. Iii and A. Singh, Eds., 2020/1// 2020, vol. 119: PMLR, pp. 8491-8501. [Online]. Available: https://proceedings.mlr.press/v119/sastry20a.html.

[12] Y. Sun, C. Guo, and Y. Li, "ReAct: Out-of-distribution Detection With Rectified Activations," in *Advances in Neural Information Processing Systems*, M. Ranzato, A. Beygelzimer, Y. Dauphin, P. S. Liang, and J. W. Vaughan, Eds., 2021, vol. 34: Curran Associates, Inc., pp. 144-157. [Online]. Available: https://proceedings.neurips.cc/paper_files/paper/2021/file/01894d6f048493d2cacde3c579c315a3-Paper.pdf.

[13] Y. Sun and Y. Li, "DICE: Leveraging Sparsification for Out-of-Distribution Detection," in *Computer Vision – ECCV 2022*, Cham, S. Avidan, G. Brostow, M. Cissé, M. Farinella Giovanni, and T. Hassner, Eds., 2022: Springer Nature Switzerland, pp. 691-708.

[14] A. Djurisic, N. Bozanic, A. Ashok, and R. Liu, "Extremely Simple Activation Shaping for Out-of-Distribution Detection," in *The Eleventh International Conference on Learning Representations*, 2023. [Online]. Available: https://openreview.net/forum?id=ndYXTEL6cZz.

[15] K. Xu, R. Chen, G. Franchi, and A. Yao, "Scaling for Training Time and Post-hoc Out-of-distribution Detection Enhancement," in *The Twelfth International Conference on Learning Representations*, 2024. [Online]. Available: https://openreview.net/forum?id=RDSTjtnqCg.

[16] D. Rezende and S. Mohamed, "Variational Inference with Normalizing Flows," in *Proceedings of the 32nd International Conference on Machine Learning*, Lille, France, F. Bach and D. Blei, Eds., 2015/1// 2015, vol. 37: PMLR, pp. 1530-1538. [Online]. Available: https://proceedings.mlr.press/v37/rezende15.html.

[17] P. Kirichenko, P. Izmailov, and A. G. Wilson, "Why Normalizing Flows Fail to Detect Out-of-Distribution Data," in *Advances in Neural Information Processing Systems*, H. Larochelle, M. Ranzato, R. Hadsell, M. F. Balcan, and H. Lin, Eds., 2020, vol. 33: Curran Associates, Inc., pp. 20578-20589. [Online]. Available: https://proceedings.neurips.cc/paper_files/paper/2020/file/ecb9fe2fbb99c31f567e9823e884dbec-Paper.pdf.

[18] L. Zhang, M. Goldstein, and R. Ranganath, "Understanding Failures in Out-of-Distribution Detection with Deep Generative Models," in *Proceedings of the 38th International Conference on Machine Learning*, Proceedings of Machine Learning Research, M. Marina and Z. Tong, Eds., 2021, vol. 139: PMLR, pp. 12427--12436. [Online]. Available: https://proceedings.mlr.press/v139/zhang21g.html.

[19] H. Kamkari, B. L. Ross, J. C. Cresswell, A. L. Caterini, R. Krishnan, and G. Loaiza-Ganem, "A Geometric Explanation of the Likelihood OOD Detection Paradox," in *Proceedings of the 41st International Conference on Machine Learning*, Proceedings of Machine Learning Research, S. Ruslan *et al.*, Eds., 2024, vol. 235: PMLR, pp. 22908--22935. [Online]. Available: https://proceedings.mlr.press/v235/kamkari24a.html.

[20] L. Dinh, J. Sohl-Dickstein, and S. Bengio, "Density estimation using Real NVP," in *International Conference on Learning Representations*, 2017. [Online]. Available: https://openreview.net/forum?id=HkpbnH9lx.

[21] D. P. Kingma and P. Dhariwal, "Glow: Generative Flow with Invertible 1x1 Convolutions," in *Advances in Neural Information Processing Systems*, S. Bengio, H. Wallach, H. Larochelle, K. Grauman, N. Cesa-Bianchi, and R. Garnett, Eds., 2018, vol. 31: Curran Associates, Inc. [Online]. Available: https://proceedings.neurips.cc/paper_files/paper/2018/file/d139db6a236200b21cc7f752979132d0-Paper.pdf.

[22] K. He, X. Zhang, S. Ren, and J. Sun, "Deep residual learning for image recognition," in *Proceedings of the IEEE Computer Society Conference on Computer Vision and Pattern Recognition*, 2016/12// 2016, vol. 2016-December: IEEE Computer Society, pp. 770-778, doi: 10.1109/CVPR.2016.90.

[23] K. Hara, H. Kataoka, and Y. Satoh, "Can Spatiotemporal 3D CNNs Retrace the History of 2D CNNs and ImageNet?," in *Proceedings of the IEEE Computer Society Conference on Computer Vision and Pattern Recognition*, 2018/12// 2018: IEEE Computer Society, pp. 6546-6555, doi: 10.1109/CVPR.2018.00685.

[24] J. Yang *et al.*, "OpenOOD: benchmarking generalized out-of-distribution detection," in *Proceedings of the 36th International Conference on Neural Information Processing Systems*, Red Hook, NY, USA, 2024: Curran Associates Inc.

[25] J. Zhang *et al.*, "OpenOOD v1.5: Enhanced Benchmark for Out-of-Distribution Detection," *Journal of Data-centric Machine Learning Research,* 2023/6// 2023, doi: 10.48550/arXiv.2306.09301.

[26] J. Yang, K. Zhou, Z. Liu, J. Yang, K. Zhou, and Z. Liu, "Full-Spectrum Out-of-Distribution Detection," *International Journal of Computer Vision 2023 131:10,* vol. 131, no. 10, 2023-06-13, doi: 10.1007/s11263-023-01811-z.

[27] J. Yang *et al.*, "MedMNIST v2 - A large-scale lightweight benchmark for 2D and 3D biomedical image classification," *Scientific Data 2023 10:1,* vol. 10, no. 1, pp. 1-10, 2023/1// 2023, doi: 10.1038/s41597-022-01721-8.

[28] B. H. Menze *et al.*, "The Multimodal Brain Tumor Image Segmentation Benchmark (BRATS)," *IEEE transactions on medical imaging,* vol. 34, no. 10, pp. 1993-2024, 2015/10// 2015, doi: 10.1109/TMI.2014.2377694.

[29] S. Bakas *et al.*, "Advancing The Cancer Genome Atlas glioma MRI collections with expert segmentation labels and radiomic features," *Scientific data,* vol. 4, 2017/9// 2017, doi: 10.1038/SDATA.2017.117.

[30] S. Bakas *et al.*, "Identifying the Best Machine Learning Algorithms for Brain Tumor Segmentation, Progression Assessment, and Overall Survival Prediction in the BRATS Challenge," *Sandra Gonzlez-Vill,* vol. 124, 2018/11// 2018, doi: 10.48550/arXiv.1811.02629.

[31] B. Lambert, F. Forbes, S. Doyle, and M. Dojat, "Multi-layer Aggregation as a Key to Feature-Based OOD Detection," pp. 104-114, 2023, doi: 10.1007/978-3-031-44336-7_11.

[32] B. Lambert, "Quantifying and understanding uncertainty in deep-learning-based medical image segmentation," Université Grenoble Alpes [2020-....], 2024. [Online]. Available: https://theses.hal.science/tel-04673383

[33] F. Isensee *et al.*, "Automated brain extraction of multisequence MRI using artificial neural networks," *Human Brain Mapping,* vol. 40, no. 17, pp. 4952-4964, 2019/12// 2019, doi: 10.1002/HBM.24750.

[34] T. Rohlfing, N. M. Zahr, E. V. Sullivan, and A. Pfefferbaum, "The SRI24 multichannel atlas of normal adult human brain structure," *Human Brain Mapping,* vol. 31, no. 5, pp. 798-819, 2010/5// 2010, doi: 10.1002/HBM.20906.

[35] N. J. Tustison *et al.*, "The ANTsX ecosystem for quantitative biological and medical imaging," *Scientific Reports,* vol. 11, no. 1, pp. 9068-9068, 2021/4// 2021, doi: 10.1038/s41598-021-87564-6.

[36] Y. Suter *et al.*, "The LUMIERE dataset: Longitudinal Glioblastoma MRI with expert RANO evaluation," *Scientific Data,* vol. 9, no. 1, 2022/12// 2022, doi: 10.1038/S41597-022-01881-7.

[37] A. F. Kazerooni *et al.*, "The Brain Tumor Segmentation (BraTS) Challenge 2023: Focus on Pediatrics (CBTN-CONNECT-DIPGR-ASNR-MICCAI BraTS-PEDs)," *ArXiv,* pp. arXiv:2305.17033v7-arXiv:2305.17033v7, 2023/5// 2023, doi: 10.48550/arXiv.2305.17033.

[38] M. Adewole *et al.*, "The Brain Tumor Segmentation (BraTS) Challenge 2023: Glioma Segmentation in Sub-Saharan Africa Patient Population (BraTS-Africa)," *ArXiv,* pp. arXiv:2305.19369v1-arXiv:2305.19369v1, 2023/5// 2023, doi: 10.48550/arXiv.2305.19369.

[39] S. Chilamkurthy *et al.*, "Development and Validation of Deep Learning Algorithms for Detection of Critical Findings in Head CT Scans," 2018/3// 2018, doi: 10.48550/arXiv.1803.05854.

[40] H. J. Kuijf *et al.*, "Standardized Assessment of Automatic Segmentation of White Matter Hyperintensities and Results of the WMH Segmentation Challenge," *IEEE Transactions on Medical Imaging,* vol. 38, no. 11, pp. 2556-2568, 2019/11// 2019, doi: 10.1109/TMI.2019.2905770.

[41] S. L. Liew *et al.*, "A large, curated, open-source stroke neuroimaging dataset to improve lesion segmentation algorithms," *Scientific Data 2022 9:1,* vol. 9, no. 1, pp. 1-12, 2022/6// 2022, doi: 10.1038/s41597-022-01401-7.

[42] F. Pérez-García, R. Rodionov, A. Alim-Marvasti, R. Sparks, J. S. Duncan, and S. Ourselin, "Simulation of Brain Resection for Cavity Segmentation Using Self-supervised and Semi-supervised Learning," *Lecture Notes in Computer Science (including subseries Lecture Notes in Artificial Intelligence and Lecture Notes in Bioinformatics),* vol. 12263 LNCS, pp. 115-125, 2020, doi: 10.1007/978-3-030-59716-0_12.

[43] "IXI Dataset – Brain Development." https://brain-development.org/ixi-dataset/ (accessed.

[44] A. E. Kavur *et al.*, "CHAOS Challenge - combined (CT-MR) healthy abdominal organ segmentation," *Medical Image Analysis,* vol. 69, pp. 101950-101950, 2021/4// 2021, doi: 10.1016/J.MEDIA.2020.101950.

[45] F. Natalia *et al.*, "Development of Ground Truth Data for Automatic Lumbar Spine MRI Image Segmentation," *Proceedings - 20th International Conference on High Performance Computing and Communications, 16th International Conference on Smart City and 4th International Conference on Data Science and Systems, HPCC/SmartCity/DSS 2018,* pp. 1449-1454, 2019/1// 2019, doi: 10.1109/HPCC/SMARTCITY/DSS.2018.00239.

[46] X. Xu, F. Zhou, B. Liu, D. Fu, and X. Bai, "Efficient Multiple Organ Localization in CT Image Using 3D Region Proposal Network," *IEEE Transactions on Medical Imaging,* vol. 38, no. 8, pp. 1885-1898, 2019/8// 2019, doi: 10.1109/TMI.2019.2894854.

[47] P. Bilic *et al.*, "The Liver Tumor Segmentation Benchmark (LiTS)," *Medical Image Analysis,* vol. 84, 2023/02/01, doi: 10.1016/j.media.2022.102680.

[48] X. Wang, Y. Peng, L. Lu, Z. Lu, M. Bagheri, and R. M. Summers, "ChestX-Ray8: Hospital-Scale Chest X-Ray Database and Benchmarks on Weakly-Supervised Classification and Localization of Common Thorax Diseases," in *2017 IEEE Conference on Computer Vision and Pattern Recognition (CVPR)*, 2017/7// 2017, vol. 2017-January: IEEE Computer Society, pp. 3462-3471, doi: 10.1109/CVPR.2017.369.

[49] D. S. Kermany *et al.*, "Identifying Medical Diagnoses and Treatable Diseases by Image-Based Deep Learning," *Cell,* vol. 172, no. 5, pp. 1122-1131.e9, 2018/2// 2018, doi: 10.1016/J.CELL.2018.02.010.

[50] J. N. Kather *et al.*, "Predicting survival from colorectal cancer histology slides using deep learning: A retrospective multicenter study," *PLOS Medicine,* vol. 16, no. 1, pp. e1002730-e1002730, 2019, doi: 10.1371/JOURNAL.PMED.1002730.

[51] P. Tschandl, C. Rosendahl, and H. Kittler, "The HAM10000 dataset, a large collection of multi-source dermatoscopic images of common pigmented skin lesions," *Scientific Data 2018 5:1,* vol. 5, no. 1, pp. 1-9, 2018/8// 2018, doi: 10.1038/sdata.2018.161.

[52] R. Liu *et al.*, "DeepDRiD: Diabetic Retinopathy—Grading and Image Quality Estimation Challenge," *Patterns,* vol. 3, no. 6, pp. 100512-100512, 2022/6// 2022, doi: 10.1016/J.PATTER.2022.100512.

[53] A. Acevedo, A. Merino, S. Alférez, Á. Molina, L. Boldú, and J. Rodellar, "A dataset of microscopic peripheral blood cell images for development of automatic recognition systems," *Data in Brief,* vol. 30, pp. 105474-105474, 2020/6// 2020, doi: 10.1016/J.DIB.2020.105474.

[54] D. P. Kingma and M. Welling, "Auto-Encoding Variational Bayes," 2013/12/20, doi: 10.48550/arXiv.1312.6114.

[55] R. T. Q. Chen, J. Behrmann, D. K. Duvenaud, and J.-H. Jacobsen, "Residual Flows for Invertible Generative Modeling," presented at the Advances in Neural Information Processing Systems, 2019, 2019. [Online]. Available: https://proceedings.neurips.cc/paper_files/paper/2019/file/5d0d5594d24f0f955548f0fc0ff83d10-Paper.pdf.

[56] C. Durkan, A. Bekasov, I. Murray, and G. Papamakarios, "Neural Spline Flows," presented at the Advances in Neural Information Processing Systems, 2019, 2019. [Online]. Available: https://proceedings.neurips.cc/paper_files/paper/2019/file/7ac71d433f282034e088473244df8c02-Paper.pdf.

[57] G. Huang, Z. Liu, L. V. D. Maaten, and K. Q. Weinberger, "Densely Connected Convolutional Networks," in *2017 IEEE Conference on Computer Vision and Pattern Recognition (CVPR)*, 21-26 July 2017 2017, pp. 2261-2269, doi: 10.1109/CVPR.2017.243.

[58] Z. Liu *et al.*, "Swin Transformer: Hierarchical Vision Transformer Using Shifted Windows," presented at the Proceedings of the IEEE/CVF International Conference on Computer Vision (ICCV), 2021/10, 2021.

[59] E. Nalisnick, A. Matsukawa, Y. W. Teh, and B. Lakshminarayanan, "Detecting Out-of-Distribution Inputs to Deep Generative Models Using Typicality," 2019/06/07, doi: 10.48550/arXiv.1906.02994.